\documentclass[a4paper,11pt]{article}

\usepackage[utf8]{inputenc}
\usepackage[T1]{fontenc}
\usepackage{geometry}
\usepackage{amsmath, amssymb, amsfonts}
\usepackage{booktabs} 
\usepackage{tabularx} 
\usepackage{hyperref} 
\usepackage{natbib}   
\usepackage{graphicx}

\title{\textbf{The Unified Non-Convex Framework for Robust Causal Inference} \\ \large Overcoming the Gaussian Barrier and Optimization Fragility}

\author{
    \textbf{Eichi Uehara} \\
    \small Aflo \\                 
    \small \texttt{eichi.uehara@aflo.one}  
}
\date{\today}

\begin{document}

\maketitle

\begin{abstract}
\noindent 
This document proposes a Unified Robust Framework that re-engineers the estimation of the Average Treatment Effect on the Overlap (ATO). It synthesizes \texorpdfstring{$\gamma$}{Gamma}-Divergence for outlier robustness, Graduated Non-Convexity (GNC) for global optimization, and a "Gatekeeper" mechanism to address the impossibility of higher-order orthogonality in Gaussian regimes.
\end{abstract}

\section{Executive Summary: The Precarious State of Causal Estimation}

The contemporary enterprise of Causal Inference stands at a critical juncture, balancing precariously between the demands of high-dimensional statistical efficiency and the chaotic, uncurated reality of modern data streams. For the better part of the last decade, the dominant paradigm in this field has been Double Machine Learning (DML), a sophisticated methodological framework introduced by \citet{chernozhukov2018double}. This framework leverages the geometric concept of Neyman orthogonality to immunize estimates of treatment effects against the inevitable errors that arise during the estimation of nuisance parameters.

While DML is theoretically elegant and has revolutionized the application of machine learning to econometrics, it possesses a fundamental Achilles' heel: it relies almost exclusively on convex loss functions---typically the squared error for regression or the logistic loss for classification. As established in the foundational robustness literature by \citet{hampel1986robust}, convex loss functions are inextricably linked to unbounded influence functions. This reliance renders the entire edifice of modern causal estimation catastrophically fragile in the presence of outliers, heavy-tailed distributions, or adversarial data contamination. A single corrupted observation in a dataset of millions, if sufficiently extreme, can leverage the convexity of the loss function to arbitrarily bias the estimated treatment effect, leading to disastrous policy miscalculations and scientific errors.

This report presents a comprehensive, mathematically rigorous proposal for a Unified Robust Framework that fundamentally re-engineers the estimation of the Average Treatment Effect on the Overlap (ATO). In the high-dimensional settings that characterize modern "Big Data," the classical "positivity assumption"---which posits that every unit has a non-zero probability of receiving treatment---is frequently violated. This violation renders standard Average Treatment Effect (ATE) estimators numerically unstable due to the presence of extreme propensity scores. We argue that merely patching existing methods is insufficient to address these dual threats of contamination and lack of overlap. Instead, the field must adopt a new paradigm that embraces non-convexity to achieve outlier robustness and utilizes overlap weighting to ensure structural stability. The framework proposed herein synthesizes four distinct, high-level research streams into a cohesive engine for robust inference.

First, we advocate for \textbf{$\gamma$-Divergence Minimization}. We propose replacing the Kullback-Leibler divergence, which underpins the standard Maximum Likelihood Estimation (MLE) and corresponds to the limit as $\gamma \to 0$, with the density power divergence developed by \citet{basu1998robust}. This modification bounds the influence function, providing what the literature describes as "super-robustness" against covariate-dependent contamination. Crucially, drawing on the work of \citet{kawashima2017robust}, we derive the analytic Bias-Correction Term necessary to restore the double-robustness property that is otherwise lost when one abandons the likelihood score.

Second, we address the optimization challenges inherent in robust estimation through \textbf{Graduated Non-Convexity (GNC)}. Recognizing that robust loss functions necessarily create treacherous, non-convex optimization landscapes riddled with local minima, we integrate GNC---a global optimization strategy rooted in the duality results of \citet{black1996unification}. GNC functions by annealing the objective function from a convex approximation to the true robust target, allowing the estimator to "lock on" to the inlier structure before outliers distort the landscape.

Third, we introduce the \textbf{Adaptive Gatekeeper Mechanism}. This component is grounded in the "Gaussian Barrier" impossibility theorem derived in the context of Orthogonal Machine Learning by \citet{mackey2018orthogonal}. This mechanism dynamically assesses the residual distribution of the data. It prevents the invocation of higher-order orthogonal moments when residuals are Gaussian---where such moments are mathematically non-existent due to Stein's Lemma constraints---and activates them only when non-Gaussianity permits, thereby ensuring the validity of confidence intervals and enabling faster convergence rates where possible.

Finally, we employ \textbf{Gamma-Lasso Regularization} for nuisance parameter estimation. We replace standard $L_1$ penalization with concave regularization paths, as detailed by \citet{taddy2017onestep}. This approach bridges the gap between computational tractability and the selection consistency required for high-dimensional control, eliminating the shrinkage bias that plagues standard Lasso estimators.

This document details the mathematical derivation, theoretical justification, and strategic implications of this paradigm shift. It is intended for an audience of academic reviewers, theoretical statisticians, and chief data scientists who demand mathematical rigor and are responsible for designing high-stakes decision-making frameworks.

\section{The Fragility of the Convex Paradigm}

\subsection{The Crisis of Unbounded Influence and the Semi-Parametric Model}
The cornerstone of standard Double Machine Learning (DML) is the semi-parametric model. In this setting, the researcher seeks to estimate a low-dimensional scalar parameter of interest, $\theta_0$ (typically representing the average treatment effect), in the presence of infinite-dimensional nuisance parameters $\eta_0$, which usually consist of the outcome regression function and the propensity score function. The canonical partially linear model is defined as follows:
\begin{align}
Y &= D\theta_0 + g_0(X) + \zeta, \quad \mathbb{E}[\zeta | X, D] = 0 \\
D &= m_0(X) + \upsilon, \quad \mathbb{E}[\upsilon | X] = 0
\end{align}
Here, $Y$ is the outcome, $D$ is the treatment indicator (or continuous dosage), and $X$ represents a high-dimensional vector of confounding covariates. The functions $g_0(X)$ and $m_0(X)$ represent the complex, potentially non-linear relationships between the covariates and the outcome and treatment, respectively. Standard estimation techniques in this domain rely heavily on minimizing the empirical risk associated with a convex loss function. For regression problems, this is typically the squared loss $\ell(y, \hat{y}) = (y - \hat{y})^2$.

The derivative of the loss function constitutes the influence function (IF), a concept central to the theory of robust statistics as formalized by \citet{hampel1986robust}. The IF dictates how the estimator responds to infinitesimal perturbations in the data distribution. For the squared loss used in standard DML, the influence function is linear: $IF(z; \theta) \propto (y - x^T\theta)x$. The criticality of this linearity cannot be overstated. It implies that as the residual $(y - x^T\theta)$ grows---as is the case with outliers or heavy-tailed errors---the influence of that single data point on the final estimate grows proportionally and without bound.

In the high-stakes environments where causal inference is most valuable, data is rarely pristine. "Contamination" occurs when a fraction of the data is generated by a mechanism distinct from the model of interest. Under standard DML, even a minute fraction of contamination (e.g., less than 1\%) can skew the estimate $\hat{\theta}$ arbitrarily far from the true $\theta_0$ if the outliers are sufficiently extreme. This fragility is not merely a theoretical curiosity; it is a practical liability. Adversarial actors in a digital economy, for instance, could inject poisoned data points to manipulate the pricing algorithms described by \citet{mackey2018orthogonal}, thereby warping the estimated demand elasticity and causing optimal pricing strategies to collapse.

Furthermore, in high-dimensional regimes, propensity scores often asymptotically approach 0 or 1, a phenomenon known as poor overlap. This causes the inverse probability weights (IPW) used in standard Average Treatment Effect (ATE) estimation to explode. \citet{li2018balancing} highlight that this amplification of variance renders the estimator unstable to "propensity outliers"---observations that are technically valid but possess extreme weights that dominate the sample. A distinct mechanism is required to handle these propensity outliers alongside traditional outcome outliers.

\subsection{The Failure of Naive Robustness and the Loss of Double Robustness}
To mitigate the sensitivity to outliers, the robust statistics literature has long suggested replacing the quadratic loss with a robust loss function $\rho(\cdot)$ that grows sub-linearly (e.g., the Huber loss) or is bounded (e.g., Tukey’s Biweight). However, simply plugging a robust loss function into the DML framework is mathematically perilous and often counterproductive.

The central issue, identified and rigorously analyzed in the recent work of \citet{kawashima2017robust}, is that robust M-estimators generally lose the property of Double Robustness (DR). A standard DR estimator remains consistent if either the propensity score model or the outcome model is correctly specified, providing a crucial safety net for the researcher. However, when one naively applies a robust weight function $w(z)$ to the estimating equations to dampen the effect of outliers, the resulting estimator requires both models to be correctly specified for the bias to vanish asymptotically. The "insurance policy" of double robustness is voided by the very mechanism intended to provide safety against outliers.

Moreover, the outlier resistance of traditional median-based methods (like $L_1$ minimization) is limited in this context. While the median has a high breakdown point of 50\%, it relies on a loss function whose derivative (the sign function) is not "redescending." This means that the influence of an outlier remains constant rather than vanishing as the outlier moves further away from the model center. For scenarios involving "covariate-dependent contamination"---where the probability of being an outlier depends on the location in the feature space $X$---median-based methods fail to recover the true causal parameter. This limitation necessitates the adoption of a new class of estimators based on Density Power Divergence, as explored by \citet{basu1998robust} and further refined for regression by \citet{kawashima2017robust}.

\subsection{The Optimization Trap: Convexity vs. Robustness}
There exists a fundamental, almost thermodynamic trade-off in estimation theory between convexity and robustness. Convexity guarantees a unique global minimum and allows for the use of efficient, gradient-based optimization algorithms. However, a theorem of robust statistics states that any convex loss function must have an unbounded influence function. Conversely, to achieve bounded influence and suppress the effect of gross outliers, the loss function must be non-convex (e.g., re-descending).

A non-convex loss landscape is treacherous. It is riddled with local minima. Standard gradient descent or Newton-Raphson algorithms are "greedy"---they descend the nearest slope to the closest minimum. In a robust estimation scenario, if the optimizer is initialized near a cluster of outliers, it will converge to a local minimum that treats the outliers as the "true" structure and the actual data as noise. This phenomenon leads to estimator inconsistency that is undetectable by standard convergence diagnostics; the algorithm reports convergence, but it has converged to the wrong model.

Existing solutions to this global optimization problem in robust statistics, such as RANSAC (Random Sample Consensus), are inherently combinatorial. RANSAC relies on randomly sampling minimal subsets of data to hypothesize a model, and then verifying that model against the full dataset. While effective in low-dimensional computer vision tasks (as noted in the work of \citet{black1996unification}), RANSAC's computational complexity explodes exponentially with the dimensionality of the parameter space. In the high-dimensional settings typical of modern causal inference---where $X$ may have hundreds or thousands of dimensions---RANSAC is computationally intractable.

Therefore, the field faces a trilemma:
\begin{enumerate}
    \item \textbf{Standard DML:} Efficient and Convex, but Fragile to Outliers.
    \item \textbf{Naive Robust M-Estimation:} Robust to outliers, but Non-Convex (trap-prone) and Biased (loss of Double Robustness).
    \item \textbf{RANSAC-style Search:} Robust and Global, but Computationally Impossible in High Dimensions.
\end{enumerate}

The Unified Framework proposed herein resolves this trilemma by leveraging Graduated Non-Convexity (GNC) to convexify the landscape dynamically, combined with an analytical Bias Correction to restore double robustness.

\section{Theoretical Framework: The Unified \texorpdfstring{$\gamma$}{gamma}-Robust Approach}
We propose the construction of a unified estimator $\hat{\theta}_{Robust}$ derived from the minimization of $\gamma$-divergence, corrected for bias, and optimized via a graduated homotopy method. This approach synthesizes the robustness of density power divergence with the structural stability of overlap weighting.

\subsection{Core Innovation I: \texorpdfstring{$\gamma$}{Gamma}-Divergence and Analytic Bias Correction}
Instead of the Kullback-Leibler divergence ($D_{KL}$), which underpins Maximum Likelihood Estimation (MLE) and corresponds to the limit $\gamma \to 0$, we adopt the Density Power Divergence (DPD) or $\gamma$-divergence. The $\gamma$-divergence was introduced by \citet{basu1998robust} as a measure of discrepancy between two probability densities that offers a smooth bridge between maximum likelihood estimation and minimum distance estimation.

For a true density $f(z)$ and a model density $g_\theta(z)$, the DPD is defined as:
\begin{equation}
d_\gamma(f, g_\theta) = \frac{1}{1+\gamma} \int f(z)^{1+\gamma} dz - \frac{1}{\gamma} \int f(z) g_\theta(z)^\gamma dz + \frac{1}{1+\gamma} \int g_\theta(z)^{1+\gamma} dz
\end{equation}
Here, $\gamma > 0$ is a hyperparameter controlling the trade-off between efficiency and robustness. When $\gamma \to 0$, this converges to the Kullback-Leibler divergence. As $\gamma$ increases, the estimator sacrifices some asymptotic efficiency at the true model for a dramatic increase in robustness to outliers. The estimating equation associated with minimizing $d_\gamma$ is the weighted score equation:
\begin{equation}
\sum_{i=1}^n w_\gamma(z_i; \theta) \cdot \psi(z_i; \theta) = 0
\end{equation}
where the weight $w_\gamma(z; \theta) = g_\theta(z)^\gamma$ naturally down-weights observations that are unlikely under the current model (i.e., outliers). Unlike the median or Huber loss, the influence function of the $\gamma$-divergence estimator is redescending: as an observation moves further away from the model center, its influence asymptotically approaches zero. This property provides superior resistance to heavy contamination.

\paragraph{Restoring Orthogonality via The Correction Term $\mathbb{B}(\eta)$}
As noted in the problem statement, naively applying this density power weight $w(Z)^\gamma$ destroys the orthogonality of the DML estimator. The weights depend on the data, creating a correlation between the weighting mechanism and the error term that introduces bias. To correct this, we must derive the Bias-Corrected Score Function $\Psi_{BC}$. Following the derivation patterns established by \citet{kawashima2017robust}, the bias introduced by the weight must be subtracted analytically.

Let the uncorrected robust score be $\Psi_{robust}(Z; \theta, \eta) = w(Z)^\gamma (Y - \mu(X, D; \theta))D$. The expectation of this score at the true parameters is generally non-zero. We define the correction term $\mathbb{B}(\eta)$ as:
\begin{equation}
\mathbb{B}(\eta) = \mathbb{E}_{f^*} [ \Psi_{robust}(Z; \theta_0, \eta_0) ]
\end{equation}
where the expectation is taken with respect to the modeled conditional density $f^*(Y|X, D)$. This expectation term calculates the "average" bias introduced by the robust weights under the assumption that the data follows the model. By subtracting this term, we re-center the score function.

The Unified Estimator targets the Average Treatment Effect on the Overlap (ATO) parameter by incorporating overlap weights $\omega(X) = e(X)(1-e(X))$ into the score function. The ATO, popularized by \citet{li2018balancing}, is the treatment effect estimated on the subpopulation where there is substantial overlap between treatment and control groups (i.e., "clinical equipoise"). It solves for $\theta$ in:
\begin{equation}
\frac{1}{n} \sum_{i=1}^n \left( \mathbf{\omega(X_i)} \cdot \left[ \Psi_{robust}(Z_i; \theta, \hat{\eta}) - \mathbb{B}(\hat{\eta}) \right] \right) = 0
\end{equation}
where $\omega(X_i) = e(X_i)(1-e(X_i))$ are the overlap weights.

\paragraph{Theoretical Implication:} This estimator employs a "Double Down-Weighting" strategy. The overlap weight $\omega(X)$ naturally suppresses observations with extreme propensity scores (propensity outliers), effectively handling the lack of positivity. Simultaneously, the density power weight $w(Z)^\gamma$ suppresses observations with large residuals (outcome outliers). The term $\mathbb{B}(\hat{\eta})$ acts as a "counter-weight" for the bias introduced by $w(Z)^\gamma$. This dual mechanism allows the estimator to remain consistent and asymptotically normal even in environments plagued by both heavy-tailed noise and poor covariate overlap, satisfying the conditions for "Robustness of type II" as discussed in the theoretical appendices of \citet{kawashima2017robust}.

\section[The Geometry of Optimization: Graduated Non-Convexity (GNC)]{The Geometry of Optimization: Graduated Non-Convexity \\ (GNC)}
The objective function implied by Equation (5) is highly non-convex due to the redescending nature of the robust weights. Standard convex solvers will fail, likely trapping the estimator in a local minimum determined by the outliers. We solve this using Graduated Non-Convexity (GNC), a method that transforms the intractable non-convex problem into a sequence of tractable approximations.

\subsection{Theoretical Justification: The Black-Rangarajan Duality}
The theoretical underpinning of GNC in this context comes from the seminal work of \citet{black1996unification} in the field of computer vision. They established a duality between robust M-estimation and "line processes." In early vision problems like edge detection, algorithms must distinguish between smooth variations in intensity (inliers) and sharp discontinuities or edges (outliers). Black and Rangarajan generalized this notion to "outlier processes."

They proved that minimizing a robust M-estimator loss $\rho(r)$ is mathematically equivalent to minimizing a joint objective over the parameters $\theta$ and a set of latent binary (or continuous) variables $w_{ij}$ called the outlier process.
\begin{equation}
\min_\theta \sum \rho(r_i(\theta)) \iff \min_{\theta, w} \sum (w_i r_i^2(\theta) + \Phi(w_i))
\end{equation}
where $\Phi(w_i)$ is a penalty function on the weights. This duality allows us to view robust estimation not just as a minimization problem, but as a selection problem: we are simultaneously selecting the model parameters and selecting which data points participate in the estimation.

\subsection{The Mechanism of GNC}
GNC operates on the principle of Homotopy Continuation. We introduce a control parameter $\mu$ (distinct from the statistical mean) that governs the "convexity" of the loss function. We construct a family of objective functions $L_\mu(\theta)$ such that:
\begin{itemize}
    \item \textbf{Convex Limit ($\mu \to \infty$):} $L_\infty(\theta)$ is strictly convex. This typically approximates the $L_2$ loss (Gaussian assumption). In the Black-Rangarajan framework, this corresponds to a state where the penalty for declaring an outlier is infinite, forcing the "line process" variables to 1 (all data is treated as inliers).
    \item \textbf{Robust Limit ($\mu \to 0$):} $L_0(\theta)$ is the target non-convex robust loss (equivalent to the $\gamma$-divergence loss). Here, the penalty for declaring an outlier is low, allowing the weights on disparate data points to drop to zero.
\end{itemize}

The algorithm proceeds as follows:
\begin{enumerate}
    \item \textbf{Initialization:} Solve $\hat{\theta}_0 = \arg\min L_{\infty}(\theta)$. Since this is convex, the global minimum is guaranteed. This solution effectively "locks on" to the bulk of the data, including outliers, but provides a geometrically centered starting point.
    \item \textbf{Graduation:} At each step $k$, reduce $\mu$ via a schedule $\mu_{k+1} = \alpha \mu_k$ (with $0 < \alpha < 1$).
    \item \textbf{Local Refinement:} Solve $\hat{\theta}_{k+1} = \arg\min L_{\mu_{k+1}}(\theta)$ using $\hat{\theta}_k$ as the initialization.
    \begin{equation}
    \hat{\theta}_{k+1} \leftarrow \text{LocalSolver}(\theta_{init} = \hat{\theta}_k, \text{Objective} = L_{\mu_{k+1}})
    \end{equation}
\end{enumerate}

By gradually evolving the landscape, GNC allows the estimator to distinguish between the "basin of attraction" of the true model and the spurious local minima created by outliers. Empirical comparisons demonstrate that GNC achieves robustness levels comparable to combinatorial methods like RANSAC (tolerating up to 70-80\% outliers) but with a computational cost closer to standard gradient descent. This bridges the gap between the intractable global search of RANSAC and the local fragility of Newton's method.

\section{Nuisance Parameter Estimation: High-Dimensional Control via Gamma-Lasso}
The estimation of nuisance parameters $\eta = (m(X), g(X))$ in high dimensions requires regularization. The standard tool for this is the Lasso ($L_1$ regularization), which is convex. However, Lasso introduces shrinkage bias: to suppress noise, it must also penalize large, true coefficients, leading to their underestimation. In the context of causal inference, underestimating a strong confounder is fatal---it leaves residual confounding that biases the treatment effect.

To address this, we employ the Gamma-Lasso (also known as Concave Regularization), derived from the work of \citet{taddy2017onestep}. The Gamma-Lasso solves:
\begin{equation}
\hat{\beta} = \arg\min_{\beta} \left( - \ell(\beta) + \sum_{j=1}^p p_\gamma(|\beta_j|) \right)
\end{equation}
where $p_\gamma$ is a concave penalty function (e.g., Minimax Concave Penalty - MCP, or SCAD).

\subsection{The One-Step Estimator Path and Diminishing Bias}
\citet{taddy2017onestep} proposes a "one-step estimator path" algorithm that allows us to traverse the regularization surface from the $L_1$ limit (pure Lasso) toward the $L_0$ limit (subset selection). This method, often referred to as POSE (Path of One-Step Estimators), adapts coefficient-specific weights to decrease as a function of the coefficient estimated in the previous path step.

The key property of these concave penalties is Diminishing Bias. The derivative of the penalty $p'_\gamma(|\beta|)$ goes to zero as $|\beta| \to \infty$. This means that once a coefficient is estimated to be sufficiently large, the penalty on it vanishes. Large, significant coefficients are not penalized in the final estimate, removing the bias inherent in Lasso.

\subsection{The Oracle Property and Practical Applications}
Under suitable conditions, the Gamma-Lasso satisfies the Oracle Property: it identifies the true support set of variables with probability approaching 1, and estimates the non-zero coefficients as efficiently as if the true sparsity pattern were known in advance. \citet{taddy2017onestep} illustrates the power of this approach with an application to evaluating the performance of hockey players---a high-dimensional problem where differentiating signal (true skill) from noise (luck/teammate effects) is notoriously difficult. In our framework, this "signal separation" capability is repurposed to separate true confounders from irrelevant covariates with high precision.

By using Gamma-Lasso for the nuisance parameters and GNC for the causal parameter, we maintain a philosophically consistent framework: Non-convexity is handled via graduated/path-based algorithms at every stage, ensuring that we reap the benefits of bounded influence and oracle selection without succumbing to optimization fragility.

\section{The "Gatekeeper" Mechanism: Adaptive Logic and the Gaussian Barrier}
A robust system must possess self-awareness of its theoretical limits. One of the most profound recent findings in Orthogonal Machine Learning is the existence of a "Gaussian Barrier" regarding higher-order orthogonality.

\subsection{The Gaussian Barrier Impossibility Theorem}
Standard DML achieves robustness to nuisance estimation errors of order $o(n^{-1/4})$ via first-order Neyman orthogonality. To handle more complex or higher-dimensional nuisances (where errors might decay slower, e.g., $o(n^{-1/6})$), researchers have attempted to construct Second-Order Orthogonal Moments.

However, \citet{mackey2018orthogonal} proved a startling impossibility theorem: If the treatment residual $\eta = D - m(X)$ follows a Gaussian distribution conditional on $X$, then no valid second-order orthogonal moments exist.

Formally, let $S$ be the set of indices for derivatives. A moment $m(Z, \theta, \eta)$ is second-order orthogonal if:
\begin{equation}
\mathbb{E} [ \nabla_\eta^2 m(Z, \theta_0, \eta_0) ] = 0
\end{equation}
The theorem leverages Stein's Lemma, which relates the expectation of a function of a Gaussian variable to its derivatives. Specifically, Stein's Lemma states that for $X \sim \mathcal{N}(0, \sigma^2)$, $\mathbb{E}[g(X)X] = \sigma^2 \mathbb{E}[g'(X)]$. Due to the recursive properties of Gaussian moments established by this lemma, the constraints required for second-order orthogonality become contradictory when the error distribution is Normal. The symmetry and specific tail decay of the Gaussian impose rigid structures on the Hessian of the score function that prevent it from vanishing.

\subsection{The Paradox of "Easy" Data}
This creates a paradox: Gaussian errors are traditionally seen as "safe" or "easy" in statistics, but in the context of higher-order debiasing, they act as a barrier. Heavy-tailed or non-Gaussian errors, while typically problematic for variance, actually enable the construction of higher-order orthogonal scores because they lack the rigid derivative constraints of the Gaussian.

We can draw a conceptual parallel to the "Gaussian barrier" in quantum physics (tunneling). In quantum mechanics, particles can tunnel through barriers that classical mechanics deems impassable. In our statistical context, the "non-Gaussianity" of the residuals is the energy that allows the estimator to "tunnel" through the $n^{-1/4}$ convergence barrier to achieve faster rates ($n^{-1/(2k+2)}$). If the data is purely Gaussian, the barrier is impenetrable.

\citet{mackey2018orthogonal} illustrate this with a pricing example. In a digital economy, if demand and price shocks are perfectly Gaussian, one is limited in how much complexity one can model in the confounding factors. If, however, the shocks are non-Gaussian (e.g., heavy-tailed demand spikes), one can actually correct for more complex confounding using higher-order orthogonality.

\subsection{The Adaptive Gatekeeper Algorithm}
To navigate the constraints imposed by the Gaussian Barrier, we introduce the Gatekeeper---a deterministic meta-algorithm that acts as a regime selector between first-order and second-order orthogonal estimation strategies.

\paragraph{Theoretical Note on Selection Consistency:}
While data-driven model selection can typically introduce post-selection inference challenges (necessitating sample splitting), our framework relies on the consistency of the distributional test. We assume the data generating process belongs to one of two distinct regimes: either Gaussian or detectably non-Gaussian. As sample size $n \to \infty$, the power of the test approaches 1 for non-Gaussian distributions, rendering the mode selection asymptotically deterministic. Therefore, we treat the selected mode as fixed for the purpose of subsequent inference, avoiding the efficiency loss associated with sample splitting.

\paragraph{Implementation Logic:}
The algorithm proceeds in a strict, sequential manner to ensure reproducibility:

\begin{enumerate}
    \item \textbf{Residual Extraction:} Compute the residuals $\hat{\nu}_i = D_i - \hat{m}(X_i)$ using the Gamma-Lasso estimates derived in Section 5.

    \item \textbf{Distributional Test (Jarque-Bera):} Apply the Jarque-Bera test to the residuals $\hat{\nu}$ to simultaneously test for skewness ($S$) and kurtosis ($K$). The test statistic $JB$ is defined as:
    \begin{equation}
    JB = \frac{n}{6} \left( S^2 + \frac{1}{4}(K - 3)^2 \right)
    \end{equation}
    Compute the p-value ($p_{val}$) under the null hypothesis of Normality ($JB \sim \chi^2_2$).

    \item \textbf{Deterministic Mode Switching:} Define a strict significance threshold $\alpha = 0.05$.

    \begin{itemize}
        \item \textbf{IF $p_{val} > \alpha$ (Mode A: Gaussian Regime):}
        \begin{itemize}
            \item \textbf{Diagnosis:} The residuals are indistinguishable from Gaussian noise. The "Gaussian Barrier" is active.
            \item \textbf{Action:} Enforce First-Order Orthogonality.
            \item \textbf{Hyperparameter:} Set the Gamma-Lasso sparsity parameter $\lambda$ to a tighter threshold (e.g., via 1-SE rule) to prioritize bias reduction over variance, targeting $o(n^{-1/4})$ convergence.
            \item \textbf{Score:} $\Psi_{final} = \Psi_{DML1}$
        \end{itemize}

        \vspace{0.2cm}

        \item \textbf{IF $p_{val} \leq \alpha$ (Mode B: Non-Gaussian Regime):}
        \begin{itemize}
            \item \textbf{Diagnosis:} Significant non-Gaussianity (heavy tails or asymmetry) is detected. The "Barrier" is permeable.
            \item \textbf{Action:} Activate Second-Order Orthogonal Moments.
            \item \textbf{Hyperparameter:} Relax the Gamma-Lasso sparsity parameter (e.g., via Min-CV rule) to allow for richer nuisance modeling ($o(n^{-1/6})$), relying on the higher-order score to debias the additional variance.
            \item \textbf{Score:} $\Psi_{final} = \Psi_{DML2}$
        \end{itemize}
    \end{itemize}
\end{enumerate}

\section{Comparison of Approaches}
The following table summarizes the key distinctions between the Unified Framework and existing paradigms.

\begin{table}[h]
\centering
\small
\caption{Comparison of Approaches}
\label{tab:comparison}
\begin{tabularx}{\textwidth}{@{} l >{\raggedright\arraybackslash}X >{\raggedright\arraybackslash}X >{\raggedright\arraybackslash}X @{}}
\toprule
\textbf{Feature} & \textbf{Standard DML} & \textbf{Naive Robust DML} & \textbf{Unified Framework} \\
 & \scriptsize \citep{chernozhukov2018double} & & \scriptsize (Proposed) \\
\midrule
\textbf{Loss Function} & Convex (Squared/Logistic) & Robust (Huber/Tukey) & Non-Convex ($\gamma$-Divergence) \\
\addlinespace
\textbf{Influence Function} & Unbounded (Linear) & Bounded (but biased) & Redescending \& Bias-Corrected \\
\addlinespace
\textbf{Optimization} & Convex (Global Min) & Non-Convex (Local Min) & Graduated Non-Convexity (GNC) \\
\addlinespace
\textbf{Nuisance Reg.} & $L_1$ Lasso (Shrinkage) & $L_1$ Lasso & Gamma-Lasso (Oracle Property) \\
\addlinespace
\textbf{Target Param.} & ATE (Unstable weights) & ATE & ATO (Overlap Weights) \\
\addlinespace
\textbf{Orthogonality} & Fixed 1st Order & Broken by robust weights & Adaptive (Gatekeeper Mechanism) \\
\bottomrule
\end{tabularx}
\end{table}

\section{Conclusion \& Implications}
The framework presented in this report---The Unified Non-Convex Framework---represents a paradigmatic shift in the estimation of causal effects. We have argued that the traditional reliance on convexity, while computationally convenient, is statistically dangerous in the modern era of high-dimensional, contaminated data.

By synthesizing $\gamma$-Divergence for robust scoring, Graduated Non-Convexity for global optimization, and Gamma-Lasso for sparse nuisance estimation, we have constructed a system that is theoretically robust to both data outliers and optimization traps. Furthermore, the Gatekeeper Mechanism ensures that this machinery respects the fundamental information-theoretic limits imposed by the Gaussian Barrier.

\paragraph{Strategic Implications for Policy and Science:}
\begin{itemize}
    \item \textbf{Resilience to Adversarial Data:} In fields like fraud detection or programmatic advertising, where agents may actively inject noise to obfuscate causal signals, the $\gamma$-robust component provides a defense mechanism that standard regression lacks.
    \item \textbf{Automated Reliability:} The GNC and Gatekeeper algorithms remove the need for manual "tuning" or "initial guessing" by the data scientist. The solver finds the global basin of attraction automatically, and the orthogonality order adjusts itself to the data distribution. This allows for the deployment of Causal AI pipelines that are safe to run without constant human supervision.
    \item \textbf{Breaking the Precision-Robustness Trade-off:} Historically, one had to choose between precise (efficient) estimators that were fragile, or robust estimators that were inefficient. The integration of density power divergence with bias correction allows us to achieve near-parametric efficiency at the model center while maintaining a high breakdown point at the tails.
\end{itemize}

This framework suggests that the future of Causal Inference lies not in simplifying our models to fit convex solvers, but in upgrading our solvers to handle the complex, non-convex reality of the world.

\section{Mathematical Appendix: Derivation of the GNC Update Equations}
To implement the GNC strategy for the $\gamma$-divergence objective, we utilize a Majorization-Minimization (MM) approach that allows for Iteratively Reweighted Least Squares (IRLS) updates.

At iteration $k$ of the GNC process, with current parameter $\theta^{(t)}$ and smoothness parameter $\mu_k$, we construct a quadratic surrogate function. The update equation for $\theta$ is derived by setting the gradient of the surrogate to zero.

Let $r_i(\theta) = Y_i - \mu(X_i, D_i; \theta)$ be the residual. The effective weight for the IRLS step must incorporate both the robust weight from the surrogate loss and the overlap weight determined in Section 3.

Let $v_i^{(t)} = \frac{\psi_{\mu_k}(r_i(\theta^{(t)}))}{r_i(\theta^{(t)})}$ be the robust weight component. The composite weight $w_i^{(t)}$ is defined as:
\begin{equation}
w_i^{(t)} = \mathbf{\omega(X_i)} \cdot v_i^{(t)} = \mathbf{e(X_i)(1-e(X_i))} \cdot \frac{\psi_{\mu_k}(r_i(\theta^{(t)}))}{r_i(\theta^{(t)})}
\end{equation}
where $\psi_{\mu_k}(\cdot)$ is the influence function of the surrogate loss $\rho_{\mu_k}$.

The update step becomes a weighted least squares solution:
\begin{equation}
\theta^{(t+1)} = (D^T W^{(t)} D)^{-1} D^T W^{(t)} (Y - \hat{g}(X))
\end{equation}
where $W^{(t)} = \text{diag}(w_1^{(t)}, \dots, w_n^{(t)})$.

\paragraph{The GNC Schedule:}
The "graduation" occurs in the definition of the robust component $v_i$.
\begin{itemize}
    \item When $\mu$ is large (early GNC steps), the function $\psi_{\mu}(r) \approx r$, so $v_i \approx 1$. The weights are determined primarily by overlap $\omega(X)$.
    \item As $\mu \to 0$ (late GNC steps), the function $\psi_{\mu}(r)$ becomes the redescending influence function of the $\gamma$-divergence. If $r_i$ is large (outlier), $v_i \to 0$, and thus the total weight $w_i \to 0$.
\end{itemize}
This smooth transition of the weights $W^{(t)}$ is the algorithmic manifestation of the convex-to-non-convex homotopy. It ensures that the estimator $\theta$ migrates smoothly from the geometric center of the data (minimizing $L_2$) to the dense center of the inliers (minimizing $\gamma$-divergence), while continuously maintaining stability via overlap weighting.

\bibliographystyle{apalike}
\bibliography{references}

\end{document}